\documentclass[11pt]{article}
\pdfoutput=1
\usepackage{acl2012}
\usepackage{times}
\usepackage{latexsym}
\usepackage{amsmath}
\usepackage{multirow}
\usepackage{url}
\usepackage{color}
\usepackage{breqn}
\usepackage{algorithm}
\usepackage{algpseudocode}
\usepackage{paralist}
\usepackage{graphicx}

\title{Lexicon Infused Phrase Embeddings for Named Entity Resolution}

\author{Alexandre Passos, Vineet Kumar, Andrew McCallum \\
  School of Computer Science \\
  University of Massachusetts, Amherst \\
  {\tt \{apassos,vineet,mccallum\}@cs.umass.edu}}

\date{}

\begin{document}
\maketitle
\begin{abstract}
  Most state-of-the-art approaches for named-entity recognition (NER)
  use semi supervised information in the form of word clusters and
  lexicons. Recently neural network-based language models have been
  explored, as they as a byproduct generate highly informative vector
  representations for words, known as word embeddings. In this paper
  we present two contributions: a new form of learning word embeddings
  that can leverage information from relevant lexicons to improve the
  representations, and the first system to use neural word embeddings
  to achieve state-of-the-art results on named-entity recognition in
  both CoNLL and Ontonotes NER. Our system achieves an F1 score of
  90.90 on the test set for CoNLL 2003---significantly better than any
  previous system trained on public data, and matching a system
  employing massive private industrial query-log data.
\end{abstract}

\section{Introduction}
\label{sec:introduction}

In many natural language processing tasks, such as named-entity
recognition or coreference resolution, syntax alone is not enough to
build a high performance system; some external source of information
is required. In most state-of-the-art systems for named-entity
recognition (NER) this knowledge comes in two forms: domain-specific
lexicons (lists of word types related to the desired named entity
types) and word representations (either clusterings or vectorial
representations of word types which capture some of their syntactic
and semantic behavior and allow generalizing to unseen word types).

Current state-of-the-art named entity recognition systems use Brown
clusters as the form of word representation
\cite{Ratinov:2009,Turian:2010,Miller:2004,Brown:1992}, or other
cluster-based representations computed from private data
\cite{lin2009phrase}. While very attractive due to their simplicity,
generality, and hierarchical structure, Brown clusters are limited
because the computational complexity of fitting a model scales
quadratically with the number of words in the corpus, or the number of
``base clusters'' in some efficient implementations, making it
infeasible to train it on large corpora or with millions of word
types.

Although some attempts have been made to train named-entity
recognition systems with other forms of word representations, most
notably those obtained from training neural language models
\cite{Turian:2010,Collobert:2008}, these systems have historically
underperformed simple applications of Brown clusters. A disadvantage
of neural language models is that, while they are inherently more
scalable than Brown clusters, training large neural networks is still
often expensive; for example, Turian et al \shortcite{Turian:2010}
report that some models took multiple days or weeks to produce
acceptable representations. Moreover, language embeddings learned from
neural networks tend to behave in a ``nonlinear'' fashion, as they are
trained to encourage a many-layered neural network to assign high
probability to the data. These neural networks can detect nonlinear
relationships between the embeddings, which is not possible in a
log-linear model such as a conditional random field, and therefore
limiting how much information from the embeddings can be actually
leveraged.

Recently Mikolov et al \cite{Mikolov:2013,Mikolov:2013b} proposed two
simple log-linear language models, the CBOW model and the Skip-Gram
model, that are simplifications of neural language models, and which
can be very efficiently trained on large amounts of data.  For example
it is possible to train a Skip-gram model over more than a billion tokens
with a single machine in less than half a day. These embeddings can also be
trained on phrases instead of individual word types, allowing for fine
granularity of meaning.

In this paper we make the following contributions.  (1) We show how to
extend the Skip-Gram language model by injecting supervisory training
signal from a collection of curated lexicons---effectively encouraging
training to learn similar embeddings for phrases which occur in the
same lexicons.  (2) We demonstrate that this method outperforms a
simple application of the Skip-Gram model on the semantic similarity
task on which it was originally tested.  (3) We show that a
linear-chain CRF is able to successfully use these
log-linearly-trained embeddings better than the other neural-network-trained
embeddings.  (4) We show that lexicon-infused embeddings let us easily
build a new highest-performing named entity recognition system on
CoNLL 2003 data \cite{conll2003} which is trained using only publicly
available data. (5) We also present results on the relatively
under-studied Ontonotes NER task \cite{weischedel2011ontonotes}, where
we show that our embeddings outperform Brown clusters.

\section{Background and Related Work}
\label{sec:backgr-relat-work}

\subsection{Language models and word embeddings}
\label{sec:language-models-word}

A statistical language model is a way to assign probabilities to all
possible documents in a given language. Most such models can be
classified in one of two categories: they can directly assign
probabilities to sequences of word types, such as is done in $n$-gram
models, or they can operate in a lower-dimensional latent space, to
which word types are mapped. While most state-of-the-art language
models are $n$-gram models, the representations used in models of the
latter category, henceforth referred to as ``embeddings,'' have been
found to be useful in many NLP applications which don't actually need
a language model. The underlying intuition is that when language
models compress the information about the word types in a latent space
they capture much of the commonalities and differences between word
types. Hence features extracted from these models then can generalize
better than features derived from the word types themselves.

One simple language model that discovers useful embeddings is 
known as {\it Brown clustering} \cite{Brown:1992}. A Brown clustering is a
class-based bigram model in which (1) the probability of a document
is the product of the probabilities of its bigrams, (2) the
probability of each bigram is the product of the probability of a
bigram model over latent classes and the probability of each class
generating the actual word types in the bigram, and (3) each word type
has non-zero probability only on a single class. Given a one-to-one assignment of
word types to classes, then, and a corpus of text, it is easy to
estimate these probabilities with maximum likelihood by counting the
frequencies of the different class bigrams and the frequencies of word
tokens of each type in the corpus. The Brown clustering algorithm
works by starting with an initial assignment of word types to classes
(which is usually either one unique class per type or a small number
of seed classes corresponding to the most frequent types in the
corpus), and then iteratively selecting the pair of classes to merge
that would lead to the highest post-merge log-likelihood, doing so until all
classes have been merged. This process produces a hierarchical
clustering of the word types in the corpus, and these clusterings have
been found useful in many applications
\cite{Ratinov:2009,koo2008simple,Miller:2004}. There are other similar
models of distributional clustering of English words which can be
similarly effective \cite{pereira1993distributional}.

One limitation of Brown clusters is their computational complexity, as
training takes $O(kV^2 + N)$x time to train, where $k$ is the number
of base clusters, $V$ size of vocabulary, and $N$ number of
tokens. This is infeasible for large corpora with millions of word
types.

Another family of language models that produces embeddings is the {\it
  neural language models}. Neural language models generally work by
mapping each word type to a vector in a low-dimensional vector space
and assigning probabilities to $n$-grams by processing their
embeddings in a neural network. Many different neural language models
have been proposed
\cite{Bengio:2003,Morin:2005,Bengio:2008,Mnih:2008,Collobert:2008,mikolov2010recurrent}.
While they can capture the semantics of word types, and often
generalize better than $n$-gram models in terms of perplexity,
applying them to NLP tasks has generally been less successful than
Brown clusters \cite{Turian:2010}.

Finally, there are algorithms for computing word embeddings which do
not use language models at all. A popular example is the CCA family of
word embeddings \cite{dhillon2012two,dhillon2011multi}, which work by
choosing embeddings for a word type that capture the correlations
between the embeddings of word types which occur before and after this
type.

\subsection{The Skip-gram Model}
\label{skip-gram-model}

A main limitation of neural language models is that they often have
many parameters and slow training times. To mitigate this, Mikolov et
al.~\shortcite{Mikolov:2013,Mikolov:2013b} recently proposed a family
of log-linear language models inspired by neural language models but
designed for efficiency. These models operate on the assumption that,
even though they are trained as language models, users will only look
at their embeddings, and hence all they need is to produce good
embeddings, and not high-accuracy language models.

The most successful of these models is the {\it skip-gram model}, which
computes the probability of each $n$-gram as the product of the
conditional probabilities of each context word in the $n$-gram
conditioned on its central word. For example, the probability
for the $n$-gram ``the cat ate my homework'' is represented as
$P(the|ate)P(cat|ate)P(my|ate)P(homework|ate)$.

\begin{figure}[t]
\centering
\includegraphics[width=0.5\textwidth]	{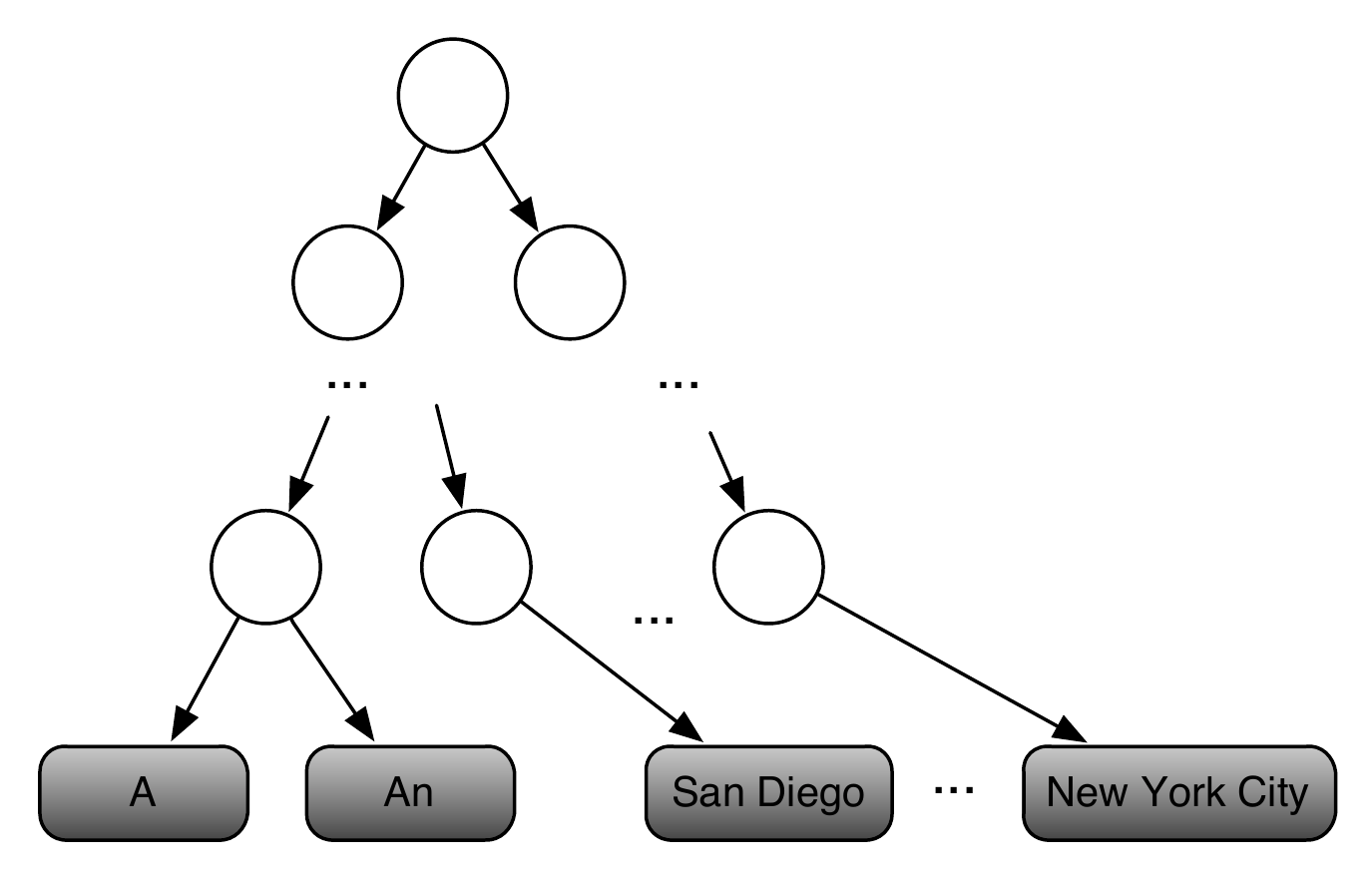}
\caption{A binary Huffman tree. Circles represent binary classifiers. Rectangles represent tokens, which can be multi-word.}
\label{HuffmanTree}
\end{figure}

To compute these conditional probabilities the model assigns an
embedding to each word type and defines a binary tree of logistic
regression classifiers with each word type as a leaf. Each classifier
takes a word embedding as input and produces a probability for a
binary decision corresponding to a branch in the tree. Each leaf in
the tree has a unique path from the root, which can be interpreted as
a set of (classifier,label) pairs. The skip-gram model then computes a
probability of a context word given a target word as the product of
the probabilities, given the target word's embeddings, of all
decisions on a path from the root to the leaf corresponding to the
context word. Figure \ref{HuffmanTree} shows such a tree structured
model.

The likelihood of the data, then, given a set $N$ of $n$-grams, with
$m_n$ being $n$-gram $n$'s middle-word, $c_n$ each context word,
$w^{c_n}_i$ the parameters of the $i$-th classifier in the path from
the root to $c_n$ in the tree, $l^{c_n}_i$ its label (either $1$ or
$-1$), $e_f$ the embedding of word type $f$, and $\sigma$ is the
logistic sigmoid function, is
\begin{equation}
  \label{eq:1}
  \prod_{n \in N} \prod_{c_n \in n} \prod_i \sigma(l^{c_n}_i {w^{c_n}_i}^T e_{m_n}).
\end{equation}

Given a tree, then, choosing embeddings $e_{m_n}$ and classifier
parameters $w^{c_n}_i$ to maximize equation \eqref{eq:1} is a
non-convex optimization problem which can be solved with stochastic
gradient descent.

The binary tree used in the model is commonly estimated by computing a
Huffman coding tree~\cite{huffman1952method} of the word types and
their frequencies. We experimented with other tree estimation schemes
but found no perceptible improvement in the quality of the embeddings.

It is possible to extend these embeddings to model phrases as well as
tokens. To do so, Mikolov et al~\shortcite{Mikolov:2013b} use a
phrase-building criterion based on the pointwise mutual information of
bigrams. They perform multiple passes over a corpus to estimate
trigrams and higher-order phrases. We instead consider candidate
trigrams for all pairs of bigrams which have a high PMI and share a
token.

\subsection{Named Entity Recognition}
\label{sec:named-entity-recogn}

Named Entity Recognition (NER) is the task of finding all instances of
explicitly named entities and their types in a given document. While
detecting named entities is superficially simple, since most sequences
of capitalized words are named entities (excluding headlines, sentence
beginnings, and a few other exceptions), finding all entities is non
trivial, and determining the correct named entity type can sometimes
be surprisingly hard.  Performing the task well often requires
external knowledge of some form.

In this paper we evaluate our system on two labeled datasets for NER:
CoNLL 2003 \cite{conll2003} and Ontonotes
\cite{weischedel2011ontonotes}. The CoNLL dataset has approximately 320k
tokens, divided into 220k tokens for training, 55k tokens for
development, and 50k tokens for testing. While the training and
development sets are quite similar, the test set is substantially
different, and performance on it depends strongly on how much external
knowledge the systems have. The CoNLL dataset has four
entity types: {\sc Person, Location, Organization, And Miscellaneous}. The
Ontonotes dataset is substantially larger: it has 1.6M tokens total,
with 1.4M for training, 100K for development, and 130k for testing. It
also has eighteen entity types, a much larger set than the
CoNLL dataset, including works of art, dates, cardinal numbers,
languages, and events.

The performance of NER systems is commonly measured in terms of
precision, recall, and F1 on the sets of entities in the ground truth
and returned by the system.

\subsubsection{Baseline System}
\label{sec:baseline-system}

In this section we describe in detail the baseline NER system we
use. It is inspired by the system described in Ratinov and Roth
\shortcite{Ratinov:2009}.

Because NER annotations are commonly not nested (for example, in the
text ``the US Army'', ``US Army'' is treated as a single entity,
instead of the location ``US'' and the organization ``US Army'') it is possible to treat NER as
a sequence labeling problem, where each token in the sentence receives
a label which depends on which entity type it belongs to and its
position in the entity. Following Ratinov and Roth
\shortcite{Ratinov:2009} we use the BILOU encoding, where each token
can either {\sc Begin} an entity, be {\sc Inside} an entity, be the
{\sc Last} token in
an entity, be {\sc Outside} an entity, or be the single {\sc Unique} token in an
entity.

Our baseline architecture is a stacked linear-chain CRF
\cite{lafferty2001conditional} system: we train two CRFs, where the
second CRF can condition on the predictions made by the first CRF as
well as features of the data. Both CRFs, following Zhang and
Johnson~\shortcite{Zhang:2003}, have roughly similar features.

While local features capture a lot of the clues used in text to
highlight named entities, they cannot necessarily disambiguate entity
types or detect named entities in special positions, such as the first
tokens in a sentence. To solve these problems most NER systems
incorporate some form of external knowledge. In our baseline system we
use lexicons of months, days, person names, companies, job titles,
places, events, organizations, books, films, and some minor
others. These lexicons were gathered from US Census data, Wikipedia
category pages, and Wikipedia redirects (and will be made publicly 
available upon publication).

Following Ratinov and Roth \shortcite{Ratinov:2009}, we also compare
the performance of our system with a system using features based on
the Brown clusters of the word types in a document. Since, as seen in
section \ref{sec:language-models-word}, Brown clusters are
hierarchical, we use features corresponding to prefixes of the path
from the root to the leaf for each word type.

More specifically, the feature templates of the baseline system are as
follows. First for each token we compute:
\begin{compactitem}
\item its word type;
\item word type, after excluding digits and lower-casing it;
\item its capitalization pattern;
\item whether it is punctuation;
\item 4-character prefixes and suffixes;
\item character $n$-grams from length 2 to 5;
\item whether it is in a wikipedia-extracted lexicon of person names
  (first, last, and honorifics), dates (months, years), place names
  (country, US state, city, place suffixes, general location words),
  organizations, and man-made things;
\item whether it is a demonym.
\end{compactitem}
For each token's label we have feature templates considering all
token's features, all neighboring token's features (up to distance 2),
and bags of words of features of tokens in a window of size 8 around
each token. We also add a feature marking whether a token is the first
occurrence of its word type in a document.

When using Brown clusters we add as token features all prefixes of
lengths 4, 6, 10, and 20, of its brown cluster.

For the second-layer model we use all these features, as well as the
label predicted for each token by the first-layer model.

As seen in the Experiments Section, our baseline system is competitive
with state-of-the-art systems which use similar forms of information.

We train this system with stochastic gradient ascent, using the
AdaGrad RDA algorithm \cite{duchi2011adaptive}, with both $\ell_1$ and
$\ell_2$ regularization, automatically tuned for each experimental
setting by measuring performance on the development set.

\subsection{NER with Phrase Embeddings}
\label{sec:ner-phrase-embeddings}

In this section we describe how to extend our baseline NER system to
use word embeddings as features.

\begin{figure}[t]
\centering
\includegraphics[width=0.5\textwidth]	{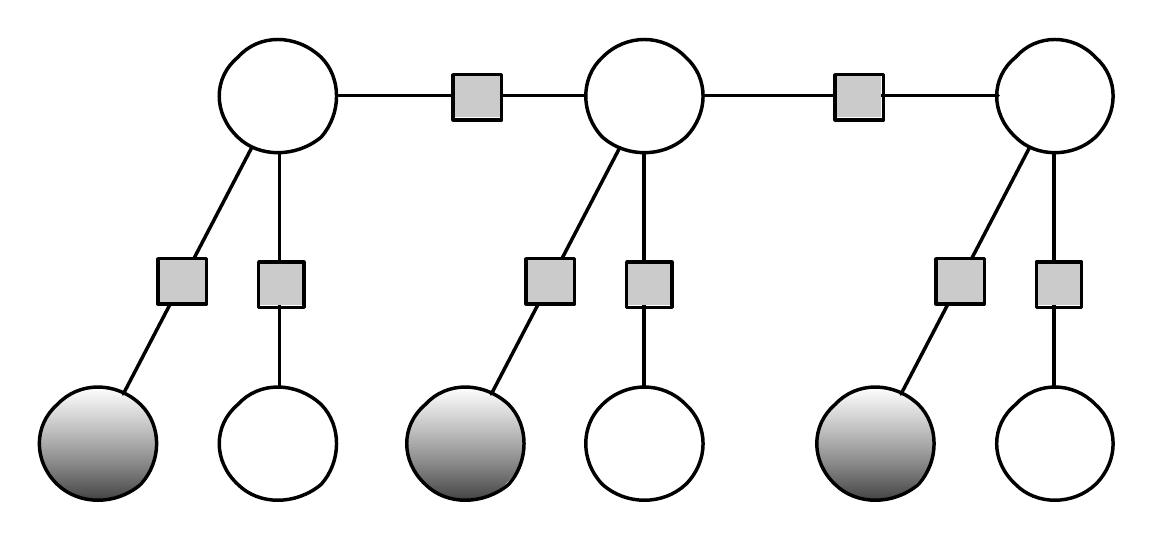}
\caption{Chain CRF model for a NER system with three tokens. Filled rectangles represent factors. Circles at top represent labels, circles at bottom represent binary token based features. Filled circles indicate the phrase embeddings for each token.}
\label{fig:nerModel}
\end{figure}

First we group the tokens into phrases, assigning to each token a
single phrase greedily. We prefer shorter phrases over longer ones,
sinceour embeddings are often more reliable for the shorter phrases,
and since the longer phrases in our dictionary are mostly extracted
from Wikipedia page titles, which are not always semantically
meaningful when seen in free text. We then add factors connecting each
token's label with the embedding for its phrase.

Figure ~\ref{fig:nerModel} shows how phrase embeddings are plugged
into a chain-CRF based NER system. Following
Turian~\shortcite{Turian:2010}, we scale the embedding vector by a
real number, which is a hyper-parameter tuned on the development
data. Connecting tokens to phrase embeddings of their neighboring
tokens did not improve performance for phrase embeddings, but it was
mildly beneficial for token embeddings.

\section{Lexicon-infused Skip-gram Models}
\label{sec:semi-supervised-skip}

The Skip-gram model as defined in Section \ref{skip-gram-model} is
fundamentally trained in unsupervised fashion using simply words and
their n-gram contexts.  Injecting some NER-specific supervision into
the embeddings can make them more relevant to the NER task.

Lexicons are a simple yet powerful way to provide task-specific
supervisory information to the model without the burden of labeling
additional data.  However, while lexicons have proven useful in
various NLP tasks, a small amount of noise in a lexicon can severely
impair the its usefulness as a feature in log-linear models.  For
example, even legitimate data, such as the Chinese last name ``He''
occurring in a lexicon of person last names, can cause the lexicon
feature to fire spuriously for many training tokens that are labeled
{\sc Person}, and then this lexicon feature may be given low or even
negative weight.

We propose to address both these problems by employing lexicons as
part of the word embedding training.  The skip-gram model can be
trained to predict not only neighboring words but also lexicon
membership of the central word (or phrase).  The resulting embedding
training will thus be somewhat supervised by tending to bring together
the vectors of words sharing a lexicon membership.  Furthermore, this
type of training can effectively ``clean'' the influence of noisy
lexicons because even if ``He'' appears in the {\sc Person} lexicon,
it will have a sufficiently different context distribution than
labeled named person entities ({\it e.g.} a lack of preceding
honorifics, etc) that the presence of this noise in the lexicon will
not be as problematic as it was previously.

Furthermore, while Skip-gram models can be trained on billions of
tokens to learn word embeddings for over a million word types in a
single day, this might not be enough data to capture reliable
embeddings of all relevant named entity phrases.  Certain sets of word
types, such as names of famous scientists, can occur infrequently
enough that the Skip-gram model will not have enough contextual
examples to learn embeddings that highlight their relevant
similarities. 

In this section we describe how to extend the Skip-gram
model to incorporate auxiliary information from lexicons, or lists of
related words, encouraging the model to assign similar embeddings to
word types in similar lexicons.

\begin{figure}[t]
\centering
\includegraphics[width=0.5\textwidth]	{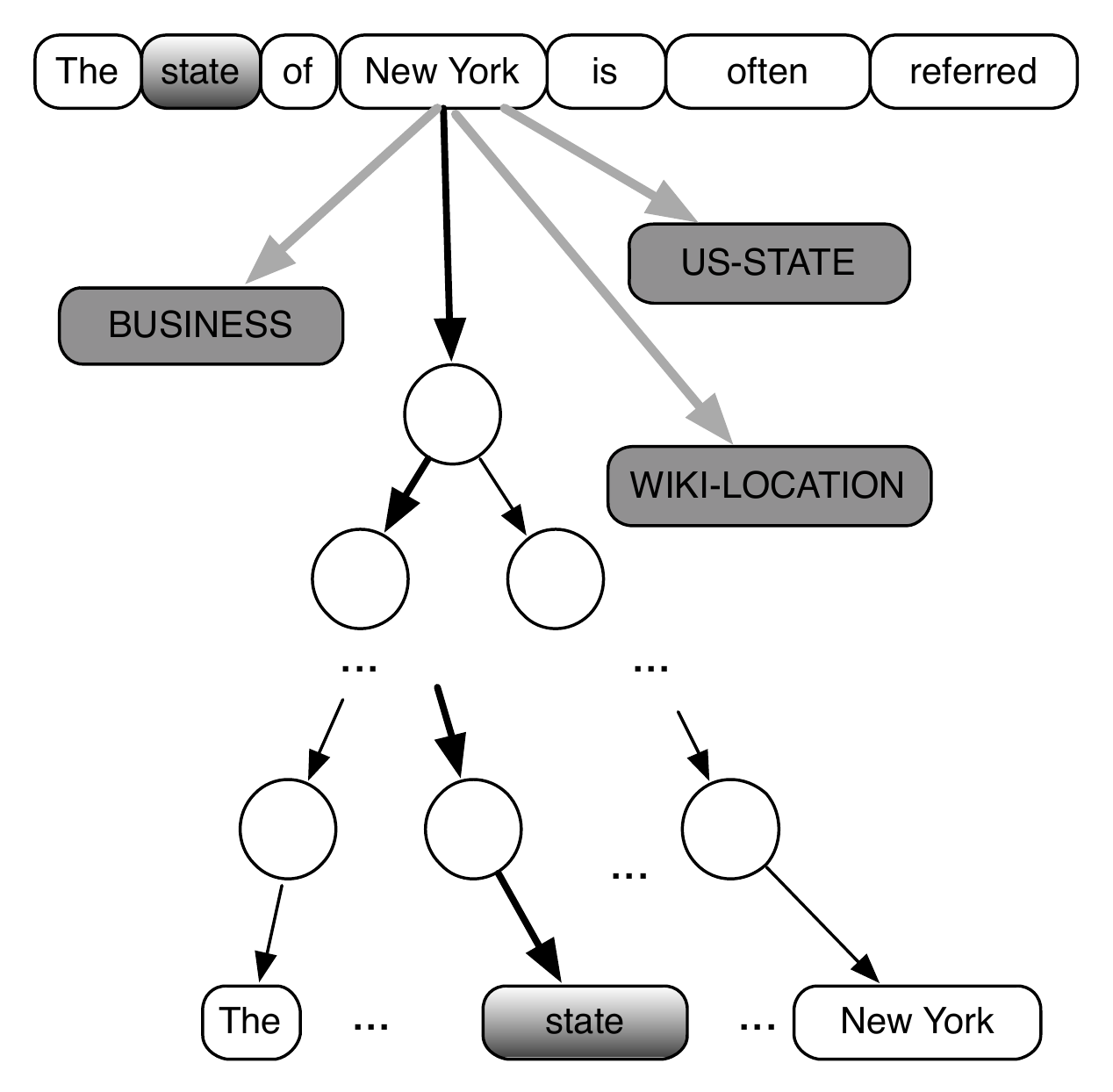}
\caption{A Semi supervised Skip-gram Model. ``New York'' predicts the word ``state''.  With lexicon-infusion, ``New York'' also predicts its lexicon classes: US-State, Wiki-location}.
\label{lexiconExample}
\end{figure}

In the basic Skip-gram model, as seen in Section
\ref{skip-gram-model}, the likelihood is, for each n-gram, a product
of the probability of the embedding associated with the middle word
conditioned on each context word. We can inject supervision in this
model by also predicting, given the embedding of the middle word,
whether it is a member of each lexicon. Figure~\ref{lexiconExample}
shows an example, where the word ``New York'' predicts ``state'', and
also its lexicon classes: Business, US-State and Wiki-Location.

Hence, with subscript $s$ iterating over each lexicon (or set of
related words), and
$l_s^{m_n}$ being
a label for whether each word is in the set, and $w_s$ indicating the parameters
of its classifier, the full likelihood of the model is
\begin{dmath}
  \label{eq:2}
  \prod_{n \in N} \left(\prod_{c_n \in n} \prod_i \sigma(l^{c_n}_i {w^{c_n}_i}^T e_{m_n})\right) \\ ~~~ \left( \prod_s \sigma(l_s^{m_n} w_s^T e_{m_n}) \right).
\end{dmath}
This is a simple modification to equation \eqref{eq:1} that also
predicts the lexicon memberships. Note that the parameters $w_s$ of
the auxiliary per-lexicon classifiers are also learned. The lexicons
are not inserted in the binary tree with the words; instead, each
lexicon gets its own binary classifier.

\begin{algorithm}[t]
\caption{Generating the training examples for lexicon-infused embeddings}
\label{trainingAlgorithm}
\begin{algorithmic}[1]
\ForAll{$n$-gram $n$ with middle word $m_n$}
  \ForAll{Context-word $c_n$}
    \ForAll{Classifier, label pair ($w^{c_n}_i$,$l^{c_n}_i$) in the tree}
      \State{Add training example $e_{m_n},w^{c_n}_i,l^{c_n}$}
    \EndFor
  \EndFor
  \ForAll{Lexicon $s$, with label $l^{m_n}_s$}
    \State{Add training example $e_{m_n},w_s,l^{m_n}_s$}
  \EndFor
\EndFor
\end{algorithmic}
\end{algorithm}

In practice, a very small fraction of words are present in a
lexicon-class and this creates skewed training data, with
overwhelmingly many negative examples. We address this issue by
aggressively sub-sampling negative training data for each lexicon
class. We do so by randomly selecting only 1\% of the possible
negative lexicons for each token.

A Skip-gram model has $V$ binary classifiers. A lexicon-infused
Skip-gram model predicts an additional $K$ classes, and thus has $V+K$
binary classifiers. If number of classes $K$ is large, we can induce a
tree over the classes, similarly to what is done over words in the
vocabulary. In our trained models, however, we have one million words
in the vocabulary and twenty-two lexicons, so this is not necessary.

\section{Experiments}
\label{sec:experiments}

Our phrase embeddings are learned on the combination of English
Wikipedia and the RCV1 Corpus~\cite{Rcv1}. Wikipedia contains 8M
articles, and RCV1 contains 946K. To get candidate phrases we first
select bigrams which have a pointwise mutual information score larger
than 1000. We discard bigrams with stopwords from a manually selected
list. If two bigrams share a token we add its corresponding trigram to
our phrase list. We further add page titles from the English Wikipedia
to the list of candidate phrases, as well as all word types. We get a
total of about 10M phrases. We restrict the vocabulary to the most
frequent 1M phrases. All our reported experiments are on
50-dimensional embeddings. Longer embeddings, while performing better
on the semantic similarity task, as seen in Mikolov et al
\shortcite{Mikolov:2013,Mikolov:2013b}, did not perform as well on
NER.

To train phrase embeddings, we use a context of length 21. We use
lexicons derived from Wikipedia categories and data from the US
Census, totaling $K=22$ lexicon classes. We use a randomly selected
0.01\% of negative training examples for lexicons.

We perform two sets of experiments. First, we validate our
lexicon-infused phrase embeddings on a semantic similarity task,
similar to Mikolov et al \cite{Mikolov:2013}. Then we evaluate their
utility on two named-entity recognition tasks.

For the NER Experiments, we use the baseline system as described in
Section~\ref{sec:baseline-system}. NER systems marked as ``Skip-gram''
consider phrase embeddings; ``LexEmb'' consider lexicon-infused
embeddings; ``Brown'' use Brown clusters, and ``Gaz'' use our
lexicons as features.

\subsection{Syntactic and Semantic Similarity}
\label{sec:synt-semant-simil}

Mikolov et al.~\shortcite{Mikolov:2013} introduce a test set to
measure syntactic and semantic regularities for words. This set
contains 8869 semantic and 10675 syntactic questions. Each question
consists of four words, such as big, biggest, small, smallest. It asks
questions of the form ``What is the word that is similar to
\textit{small} in the same sense as \textit{biggest} is similar to
\textit{big}''. To test this, we compute the vector $X =
vector(``biggest") - vector(``big") + vector(``small")$. Next, we
search for the word closest to X in terms of cosine distance
(excluding ``biggest'', ``small'', and ``big''). This question is
considered correctly answered only if the closest word found is
``smallest''. As in Mikolov et al~\cite{Mikolov:2013}, we only search
over words which are among the 30K most frequent words in the
vocabulary.

\begin{table}[t]
\begin{center}
\begin{tabular}{|l|r|}
\hline \bf Model & \bf Accuracy \\ \hline
Skip-Gram & 29.89  \\
Lex-0.05 & 30.37 \\
Lex-0.01 & \bf 30.72 \\
\hline
\end{tabular}
\end{center}
\caption{\label{skip-gram} Accuracy for Semantic-Syntactic task, when restricted to Top 30K words. Lex-0.01 refers to a model trained with lexicons, where 0.01\% of negative examples were used for training. }
\end{table}

Table~\ref{skip-gram} depicts the accuracy on Semantic Syntactic Task
for models trained with 50 dimensions. We find that lexicon-infused
embeddings perform better than Skip-gram. Further, lex-0.01 performs
the best, and we use this model for further NER experiments. There was
no perceptible difference in computation cost from learning lexicon-infused
embeddings versus learning standard Skip-gram embeddings.

\subsection {CoNLL 2003 NER}
\label{section:conll2003-results}
We applied our models on CoNLL 2003 NER data set. All hyperparameters were tuned by
training on training set, and evaluating on the development set. Then
the best hyperparameter values were trained on the combination of
training and development data and applied on the test set, to obtain
the final results. 


\begin{table}[t]
\begin{center}
\begin{tabular}{|l|l|l|}
\hline \bf System & \bf Dev & \bf Test \\ \hline
Baseline & 92.22 & 87.93\\
Baseline + Brown & 93.39 &  90.05 \\
Baseline + Skip-gram & 93.68 & 89.68 \\
Baseline + LexEmb &   93.81&  89.56\\
Baseline + Gaz & 93.69 & 89.27 \\
Baseline + Gaz + Brown & 93.88 & 90.67 \\
Baseline + Gaz + Skip-gram & 94.23 & 90.33 \\
Baseline + Gaz + LexEmb & \bf 94.46 & \bf 90.90 \\
\hline
Ando and Zhang~\shortcite{ando2005high} & 93.15 & 89.31\\
Suzuki and Isozaki~\shortcite{suzuki2008semi} & \bf 94.48 & 89.92 \\
Ratinov and Roth~\shortcite{Ratinov:2009} & 93.50 & 90.57 \\
Lin and Wu~\shortcite{lin2009phrase} & - & \bf 90.90\\
\hline
\end{tabular}
\end{center}
\caption{\label{ner:conll} Final NER F1 scores for the CoNLL 2003 shared
  task.  On the top are the systems presented in this paper, and on the bottom we have baseline systems. The best results within each area are highlighted in bold.  Lin and Wu~2009 use massive private industrial query-log data
  in training.}
\end{table}

Table \ref{ner:conll} shows the phrase F1 scores of all systems we
implemented, as well as state-of-the-art results from the
literature.  Note that using traditional unsupervised Skip-gram
embeddings is worse than Brown clusters.
In contrast, our lexicon-infused phrase embeddings \textbf{Lex-0.01}
achieves 90.90---a state-of-the-art F1 score for the test set. This result
matches the highest F1 previously reported, in Lin and Wu
\shortcite{lin2009phrase}, but is the first system to do so
without using massive private data.
Our result is signficantly better than the previous best using public
data.

\subsection{Ontonotes 5.0 NER}
\label{sec:ontonotes-5.0-ner}

Similarly to the CoNLL NER setup, we tuned the hyperparameters on the
development set. We use the same list of lexicons as for CoNLL NER.

Table~\ref{ner:ontonotes} summarize our results. We found that both
Skip-gram and Lexicon infused embeddings give better results than
using Brown Clusters as features. However, in this case Skip-gram embeddings give
marginally better results.  (So as not to jeopardize our ability to
fairly do further research on this task, we did not analyze the test
set errors that may explain this.)
These are, to the best of our knowledge, the first
published performance numbers on the Ontonotes NER task.

\begin{table}[t]
\begin{center}
\begin{tabular}{|l|l|l|}
\hline \bf System & \bf Dev & \bf Test \\ \hline
Baseline & 79.04 & 79.85\\
Baseline + Brown & 79.95 & 81.38 \\
Baseline + Skip-gram & 80.59 &  81.91 \\
Baseline + LexEmbd &  80.65& 81.82\\
Baseline + Gaz & 79.85 & 81.31 \\
Baseline + Gaz + Brown & 80.53 & 82.05 \\
Baseline + Gaz + Skip-gram &80.70  & \bf 82.30\\
Baseline + Gaz + LexEmb & \bf 80.81 & 82.24 \\
\hline
\end{tabular}
\end{center}
\caption{\label{ner:ontonotes} Final NER F1 scores for Ontonotes 5.0 dataset. The results in bold face are the best on each evaluation set.}
\end{table}

\section{Conclusions}
We have shown how to inject external supervision to a
Skip-gram model to learn better phrase embeddings. We demonstrate the
quality of phrase embeddings on three tasks: Syntactic-semantic
similarity, CoNLL 2003 NER, and Ontonotes 5.0 NER. In the process,
we provide a new public state-of-the-art NER system for the widely
contested CoNLL 2003 shared task. 

We demonstrate how we can plug phrase embeddings into an existing
log-linear CRF System. 

This work demonstrates that it is possible to learn high-quality
phrase embeddings and fine-tune them with external supervision from
billions of tokens within one day computation time. We further demonstrate that learning
embeddings is important and key to improve NLP Tasks such as NER.

In future, we want to explore employing embeddings to other NLP tasks
such as dependency parsing and coreference resolution. We also want to
explore improving embeddings using error gradients from NER.

\bibliography{refs}
\bibliographystyle{acl2012}
\end{document}